%% file: main.tex
\def\year{2022}\relax
\documentclass[letterpaper]{article} 
\usepackage{aaai22}  
\usepackage{times}  
\usepackage{helvet}  
\usepackage{courier}  
\usepackage{bbold}
\usepackage[hyphens]{url}  
\usepackage{graphicx} 
\usepackage{natbib}  
\usepackage{caption} 
\DeclareCaptionStyle{ruled}{labelfont=normalfont,labelsep=colon,strut=off} 
\usepackage{algorithm}
\usepackage{algorithmic}
\usepackage{amsmath,amsthm,amssymb}
\usepackage{xcolor}
\usepackage{newfloat}
\usepackage{listings}
\floatstyle{ruled}
\newfloat{listing}{tb}{lst}{}
\floatname{listing}{Listing}

\title{Computing Policies That Account For The Effects Of Human Agent Uncertainty During Execution In Markov Decision Processes}

\author {
    Sriram Gopalakrishnan,\textsuperscript{\rm 1}
    Mudit Verma, \textsuperscript{\rm 1}
    Subbarao Kambhampati \textsuperscript{\rm 1} \\
}
\affiliations {
    \textsuperscript{\rm 1} Arizona State University \\
     sgopal28@asu.edu
}

\begin{document}

\maketitle

\begin{abstract}

When humans are given a policy to execute, there can be policy execution errors and deviations in policy if there is uncertainty in identifying a state. This can happen due to the human agent's cognitive limitations and/or perceptual errors. So an algorithm that computes a policy for a human to execute ought to consider these effects in its computations. An optimal Markov Decision Process (MDP) policy that is poorly executed (because of a human agent) maybe much worse than another policy that is suboptimal in the MDP, but considers the human-agent's execution behavior. In this paper we consider two problems that arise from state uncertainty; these are erroneous state-inference, and extra-sensing actions that a person might take as a result of their uncertainty. We present a framework to model the human agent's behavior with respect to state uncertainty, and can be used to compute MDP policies that accounts for these problems. This is followed by a hill climbing algorithm to search for good policies given our model of the human agent. We also present a branch and bound algorithm which can find the optimal policy for such problems. We show experimental results in a Gridworld domain, and warehouse-worker domain. Finally, we present human-subject studies that support our human model assumptions. 

\end{abstract}

\section{Introduction}

Markov Decision Processes (MDPs) have been used extensively to model settings in many applications(\cite{ApplicMDP_boucherie2017markov},\cite{ApplicMDP_hu2007markov},\cite{ApplicMDP_white1993survey}) but when the agent that has to act in such a scenario is a human, we need to consider how the execution changes. In this work we focus on problems during execution by a human agent that arise from uncertainty during state inference. Specifically, we focus on two issues; erroneous state inference, and extra-sensing policy that humans take due to state uncertainty. Uncertainty can arise from perceptual limitations, and/or cognitive limitations. When a person is uncertain about the current state, we may take additional sensing actions, or take longer to decide and may still end up deciding incorrectly. For example, if we had to determine our position correctly and we were uncertain, we might repeat sensing actions to locate ourselves better. 



When we have such uncertainty, we may take additional sensing actions to try and resolve the uncertainty or repeat unhelpful previous perceptual actions. We treat these actions as coming from an extra-sensing policy of the human, as they are not part of the state-action mapping in the policy given to execute; they are a consequence of the human's uncertainty. We cannot programmatically control these, as we would with a non-human agent. In this work, we describe the extra-sensing policy as just a unique extra-sensing action per state.


\begin{figure}[H]
\centering
     \includegraphics[width=\columnwidth,scale=0.5]{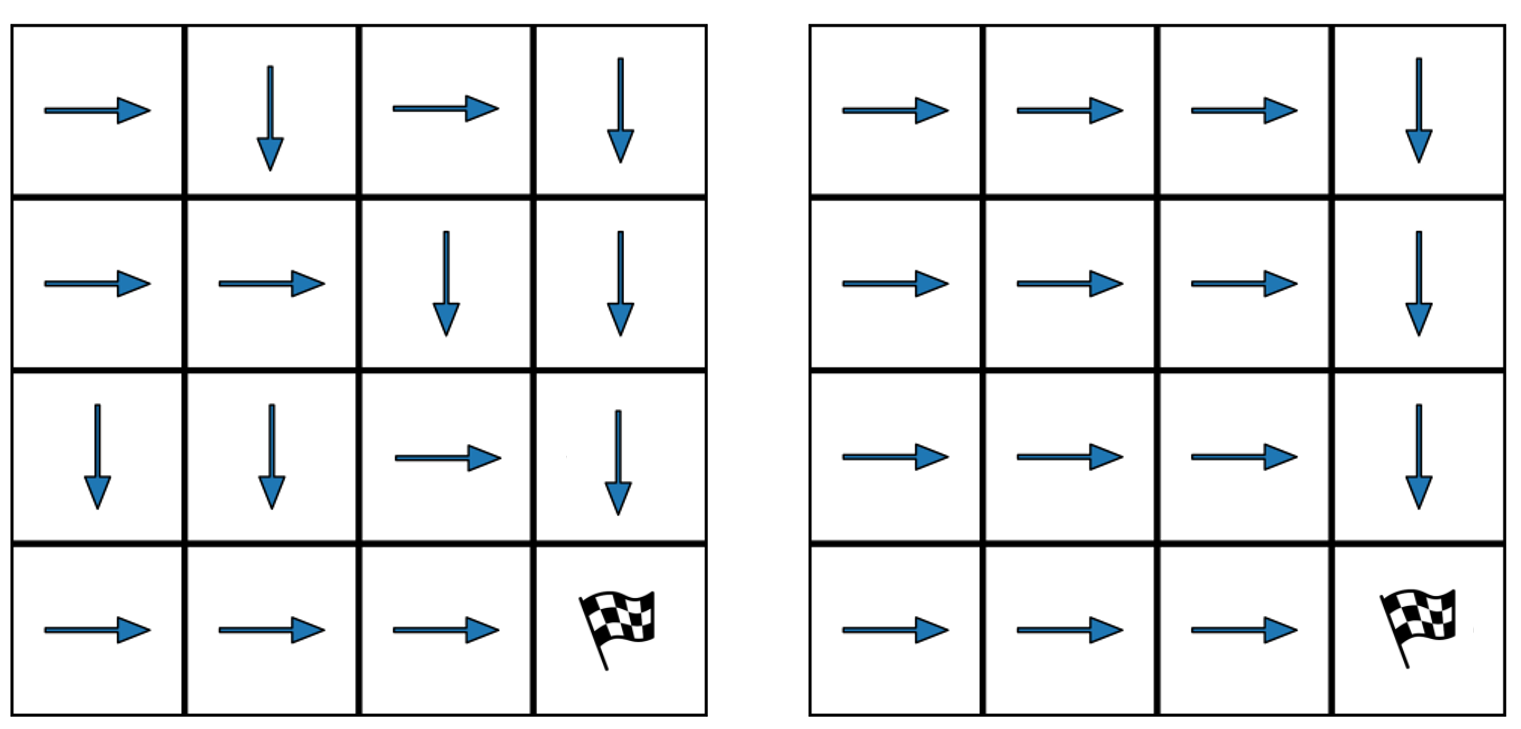}
      \caption{ Policies with identical value in an MDP, but can have different values after accounting for errors and uncertainty effects}
      \label{fig:policies_5x5}
\end{figure}

As a result of the extra-sensing action, we consider cases when the agent's uncertainty might get reduced, as well as cases where the additional sensing does not make a difference (in expectation) in identification errors; a human agent might do them anyway as they are the only options to resolve uncertainty. In our work, we are \emph{not} trying to remove uncertainty during execution; we expect it to persist in certain problems despite efforts to minimize it. Efforts to minimize it are important, but orthogonal to our work (we discuss these in the related work). We instead try to account for the effects of uncertainty when computing a policy to get better policies. An optimal policy (in terms of domain dynamics) which ignores the effects of uncertainty on execution behavior can be markedly suboptimal. To help build intuition, we present an example in the gridworld domain in Figure \ref{fig:policies_5x5}. In this toy example, the human might be uncertain as to their current position, confusing it with neighboring positions. The human's directive is to follow the given policy and take the right actions. So, when actions for neighboring positions (states) are different, they may take additional sensing actions to try and determine their position correctly. They are uncertain as to what the right action is. If the action was the same across the states they thought possible, then there is no uncertainty as to what the right action is. This let's them act right way without additional sensing actions. Figure \ref{fig:policies_5x5} shows two policies with identical policy value (by domain rewards and dynamics) and can have different values after accounting for the effects of uncertainty; in that problem the only reward is in reaching the goal position, and all actions have the same cost. The policies will have very different values if we consider that the person takes an extra sensing-action when confused between neighboring states which have different actions. Even if the perception action cost is zero, the state values computed change (assuming discounted future rewards). 

The likelihood of the human taking an extra-sensing action can be affected by not only the problem states, but also the human agent's mental state. Mental modeling a human agent can be challenging. However, we do not need a complex mental model for this problem. What we need are the likelihoods of events related to how humans might be uncertain about states, and how they respond. Using a probabilistic model of the human agent's behavior, we can compute better policies for human execution by accounting for how people make mistakes and respond to uncertainty.


In this paper, we formally define the problem of computing a policy that accounts for human uncertainty during execution in an MDP by converting it into a constrained Partially Observable Markov Decision Process (POMDP). Our contributions include : (1)A model of human agent behavior under state uncertainty (2) A Hill Climbing algorithm and branch-and-bound algorithm to find reactive controller policies that incorporate our human model and account for extra-sensing actions. (3) We show experimental results for our approach on two domains; a Gridworld domain, and a warehouse-worker domain. (4)Lastly, we present the results of a human subject study which shows evidence for the uncertainty effects on human agent execution that we model in this work.

\section{Human Model And Problem Definition}    

In this work, we assume that the underlying MDP problem is fully observable (the ground truth state is knowable), and an AI agent can detect the state and execute the optimal policy perfectly. However, when the human enacts a policy, the human's limitations leads to suboptimal execution due to state uncertainty during execution. So we need a policy that accounts for the human's state uncertainty and associated execution behavior. We will build up to the formal definition of the problem by first discussing the human model used. Then we will define how it is incorporated it into a POMDP for computing a better policy for humans.

\subsection{Human Model}


The human model is defined using the probability of inference events and extra-sensing events given a ground-truth state. For a set of domain states $S$, we define the human model as $H = \langle p_c, p_u,\psi_0,\psi_1 \rangle$, and the terms are defined as follows:

\begin{itemize}
    \item $p_c: S\times S \rightarrow [0,1]$ gives the likelihood of classifying(identifying) one state as another; $p_c(\hat{s}|s^*)$ where $s^*$ is the true state, and $\hat{s}$ is the best-guess state that the human agent thinks it is. We will use the $\hat{s}$ symbol above a state to indicate the human's guess of the state. 
    
    \item $p_u: S\times \{S_i \in 2^s\} \rightarrow [0,1]$ gives the likelihood of being uncertain between a set of states ($S_i$) for a given true state. For example, $p_u(\{s_i,s_j\}|s^*)$ is the probability of the human considering $\{s_i,s_j\}$ as the possible states when the true state is $s^*$. We will refer to such $S_i$ sets as a ``possible-set", and reflects the mental-state or belief state of the agent; it represents what states they think are possible. 
    
    \item $\psi_1: S \rightarrow [0,1]$ is a scaling-factor that affects the probability of the extra-sensing action being taken when the agent is uncertain about the right policy action. If there is more pressure to act, or the human-agent tends to be impulsive, this number would be lower to make the probability of extra-sensing action lower.
    
    \item $\psi_0: S \rightarrow [0,1]$ is a bias term that affects the probability of the extra-sensing action being taken. So even if the human infers only one state, they may not feel confident and take additional sensing actions anyway; this is what the bias term ($\psi_0$) captures.
    
\end{itemize}

\subsection{POMDP With Human Execution Under Uncertainty}
\begin{figure}[H]
\centering
     \includegraphics[width=0.7\columnwidth,height=4cm]{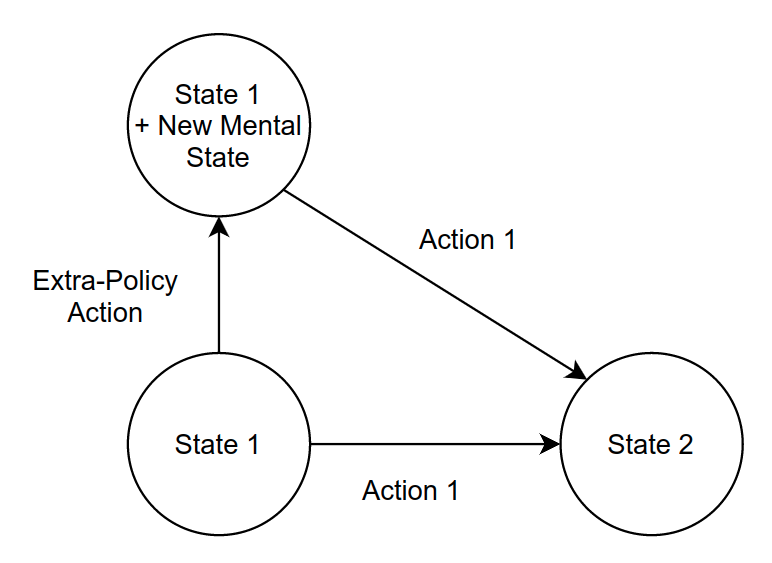}
      \caption{Additional state added to MDP for State 1 to account for different inference likelihoods by the human agent}
      \label{fig:extra_policy_mental_state}
\end{figure}

Given this human model, the problem of a POMDP with Human Execution under Uncertainty (POMDP-HUE) is defined by the tuple $\langle S,A,T,r,\gamma,p_i,H,S_2 \rangle$. Each of the terms are defined as follows: 


\begin{itemize}
    \item $S$ is the set of states in the problem.
    
    \item $A$ is the set of actions in the domain, and an additional $a^{+}$ which is the extra-sensing action.
    
    \item $S_2$ contains one successor state to every state in $S$, and is reached only when $a^{+}$ is taken. This captures the change in human's mental state (belief state) after the extra-sensing action. This is illustrated for a single state in Figure \ref{fig:extra_policy_mental_state}. This is effectively folding the belief state into the state space. 
    
    \item $T: S \times A \times S \rightarrow [0,1]$ is the transition function that outputs the likelihood of transition from one state to a successor state after an action. This includes the extra-sensing action ($a^{+})$ dynamics.
    
    \item $r: S \times A \rightarrow \mathbb{R}$ is the reward function. This includes the cost/reward associated to extra-sensing actions.
    
    \item $\gamma$ is the discount factor
    
    \item $p_i:S \rightarrow [0,1]$ is the probability of a state being the initial state
    
    \item $H$ refers to the human model as defined by $\langle p_c,p_u,\psi_1,\psi_0 \rangle$. These are separate functions for states in $S$ versus those in $S_2$, as the uncertainty can change after extra-sensing action. Both sets of functions are still defined over only the problem states $S$ as that is what the human is uncertain over. We discuss how to compute these values the section on ``Computing Human Model Parameters". 
    
\end{itemize}

The objective in the POMDP-HUE problem is to output a deterministic policy ($\pi_d:S \rightarrow A$) that optimizes for policy value (equation \ref{eqn:oneHot_policy_value}) after accounting for effects of uncertainty --that we will shortly formalize-- determined by the human model $H$. The deterministic policy is a mapping from a problem state to an action. The state as inferred by the human can be seen as a noisy observation emitted by the real state. This makes it a POMDP, and why we refer to the objective as computing a reactive controller for a POMDP. Reactive controllers were defined in influential prior works \cite{littman1994memoryless} \cite{meuleau2013solving}.

A reactive controller for a POMDP implies a control policy based on the current state's observation (human inference) only. This means that we make the assumption in this work, that the human's inference and uncertainty is predominantly influenced by the current state only. 
Furthermore, we consider that if a person is uncertain over a set of possible states in their mind, and if the policy conflicts between these states, then the human is more likely to take extra-sensing actions to try to resolve uncertainty (which we show in our human subject studies). Put another way, the uncertainty that really matters when executing a policy, is about what the right action is; it is not to perfectly detect the state. If the action is the same across states then there is no need for additional resolution. Building on this, we define the likelihood of the extra-sensing action (which is denoted by $a^{+}$) being taken as a function over the policy. 

Since extra-sensing actions and uncertainty affect execution of a policy, any deterministic policy given to the human ($\pi_d$), when actually executed by the human can become a stochastic policy; $\pi_{p}:(S \times A | \pi_d) \rightarrow [0,1]$. The first effect from uncertainty, is the likelihood of taking an extra-sensing action in a state ($s* \in S \bigcup S_2$) is defined as follows:

\begin{equation}
    \begin{split}
     \pi_{p}(s*,a^{+}|\pi_d) =\psi_0(s*) +  (1-\psi_0(s*)) \times \psi_1(s*) \times \\
     \sum_{S_i \in 2^S} p_u(S_i|s*) \times \mathbb{1} [ 0 < \sum_{s_1,s_2 \in S_i}\mathbb{1}[\pi_d(s_1) \neq \pi_d(s_2)] ]
   \end{split}
    \label{eqn:extra-sensing_action_prob}
\end{equation}

where $\mathbb{1}[.]$ is the indicator function. Succinctly, the equation says is that if one of the states in the possible-set ($S_i$) has a policy action that doesn't match with another, then the likelihood of extra-sensing action ($a^+$) can increase by $p_u(S_i|s*)$. For any given state, the number of possible states of uncertainty are likely to be few, not the full powerset ($2^S$). For example, in a gridworld setting, the current state's possible-sets may involve neighboring states (positions), but not ones further away. 

The extra-sensing action could translate to many types of actions; these could include calling a supervisor or colleague for help, or re-checking state features. The specific dynamics of the extra-sensing actions are domain dependent. For this paper's presentation, we limit the effect of extra-sensing actions in that it can improve the inference likelihood of the human agent, but doesn't change the ground truth state. In terms of the problem definition, this means the POMDP-HUE state transitions from a state in $S$ to the corresponding state in $S2$ like in Figure \ref{fig:extra_policy_mental_state}.

In addition to taking extra-sensing actions, the other effect of uncertainty is on the likelihood of choosing a policy action from the given policy $\pi_d$, and is as follows: 

\begin{equation}
    \begin{split}
    \pi_{p}(s*,a| \pi_d) = (1-\pi_{p}(s*,a^{+},\pi_d)) \times \\ 
    \sum_{s_i \in S}p_c(\hat{s_i}|s)*\mathbb{1}[\pi_d(s_i) = a]
    \end{split}
    \label{eqn:stochasticized_policy}
\end{equation}

This means that the likelihood of an action is the sum of the likelihood of it's corresponding states in the policy being inferred as the current state by the human. This is multiplied by the probability of not taking the extra-sensing action in the state, which ensures the probabilities sum to 1. 

Finally, the overall value of the \emph{determinisitic} policy $\pi_d$ given to the human is defined as:

\begin{equation}
    V(\pi_d) = \sum_{s \in S}p_i(s)*V_{\pi_{p}}(s)
    \label{eqn:oneHot_policy_value}
\end{equation}
where $V_{\pi_{p}}(s)$ is the state value in the input MDP with the stochastic policy $\pi_{p}$ which included the effects of uncertainty. The value is a weighted sum of state value, where the weights are the initial state likelihood given in $p_i(.)$.

\section{Computing Human Model Parameters}

This work focuses on the computing the policy for a human model in terms of $\langle p_c, p_u,\psi_0,\psi_1 \rangle$. Part of the appeal of this approach to modeling the human agent, is that we only need probabilities of events; this can be estimated from empirical data. We present an empirical approach to collecting the data needed to compute the human model parameters; we use this approach in our human subject studies as well. 

Since the model parameters are dependent on the state, we can collect data by testing the human agent on just the task of state detection. We would count the following with each state inference made by the human.

\begin{itemize}
    \item $C_i(\hat{s_i}|s^*)$:The number of times a state ($s_i$) was inferred for a given ground truth state ($s^*$)
    \item $C_u(S_i|s^*)$: The number of times the person was uncertain over a set of states ($S_i \subset 2^S$) for a given state.
    \item $C_{e1}(s^*)$: How often a person took an extra-sensing action (after the specified standard perceptual policy) \emph{when} the reported they were uncertain over two or more possible states.
    \item $C_{e0}(s^*)$: How often a person took an extra-sensing action when uncertain, \emph{and} their set of possible states was either 1 or none. This information helps to define the bias likelihood of taking extra actions. 
\end{itemize}

Note than when a person takes an extra-sensing action, we consider all subsequent counts separately to compute the human model for states in the set $S_2$; this is in accordance with the transition illustrated in Figure \ref{fig:extra_policy_mental_state}. 

Using this information:
\begin{equation}
    p_c(\hat{s_i}|s^*) = \frac{C_i(\hat{s_i}|s^*)}{\sum_{s_j \in S}C_i(s_j|s^*)}
\end{equation}

\begin{equation}
    p_u(S_i|s^*) = \frac{C_u(S_i|s^*)}{\sum_{S_j \in 2^S}C_u(S_j|s^*)}
\end{equation}

\begin{equation}
    \psi_1(s^*) = \frac{C_{e1}(s^*)}{\sum_{S_j \in \{S: S \in 2^S,|S|>1\}}C_u(S_j|s^*)}
\end{equation}

\begin{equation}
    \psi_0(s^*) = \frac{C_{e0}(s^*)}{\sum_{S_j \in \{S: S \in 2^S,|S|\leq 1\}}C_u(S_j|s^*)}
\end{equation}


These definitions are so that $\psi_0$ captures the baseline likelihood of taking extra-sensing action, i.e. likelihood of extra-sensing action even without any inference conflict. $\psi_1$ is the likelihood of taking extra-sensing actions when uncertain; this covers the cases when the human's inference results in 2 or more states being possible (conflicting inference). 

If the extra-sensing action sufficiently changes the likelihoods of inference, then we need to compute the model parameters for this new ``mental" state after the extra action as well. This new mental state is an additional state that is added to the MDP for each state that needs it (in our approach). It represents the updated mental state of the person.

This new mental state is reached only by the extra-sensing action from the original state. Otherwise it has the same dynamics as the preceeding state. This is illustrated in Figure \ref{fig:extra_policy_mental_state}. The state dynamics is the same as the state that preceeds it, and represents the updated mental state of the agent. There can be many such successive mental states for different extra-sensing actions taken from the same state, but we limit ourselves here to just one additional state associated to one extra-sensing action per state for this presentation. 

As for the scaling-factor $\psi_1(s)$ in the model, recall it represents how likely the human is to execute an extra-sensing when uncertain. This scaling-factor can be less than 1, as we do not expect a person to just get stuck in a state and never act unless they are completely certain. They may act regardless of uncertainty because it may --for example-- be impossible to reach a state of certainty for all state instances. One can also think of this scaling-factor as a reflection of the pressure to act on the human agent, or the patience of the agent to resolve uncertainty. 

The bias-factor $\psi_0(s)$ is the likelihood of taking an extra-sensing action. This applies to the event when the possible-set is either empty or a singleton-set. This reflects how often the person does not feel confident in their inference of a single state, or couldn't infer a state at all (made a mistake).

\section{Policy Computation For POMDP-HUE}    

Finding an optimal solution to the POMDP-HUE problem is at least as challenging as computing a reactive/memoryless controller for a POMDP, which is what our problem reduces to if one ignores the extra-sensing action (by setting $\psi_1=0 ,\psi_0=0$). Computing a reactive controller has been shown to be NP-hard (\cite{littman1994memoryless}). To handle this computational complexity, we present two algorithms. One is a hill-climbing algorithm for computing good albeit suboptimal policies quickly, and for larger state spaces. The other is a branch and bound algorithm for computing the optimal policy at higher computational cost, which is suitable for smaller state spaces and for bounding the suboptimality of the hill-climbing approach for larger state spaces.

\subsection{Human-Agent Policy Iteration(HAPI)}
We call our hill climbing approach Human-Agent Policy Iteration (HAPI) which takes the greedy best step to change the policy while accounting for uncertainty effects. In HAPI we start with a random deterministic policy ($\pi_d$), and compute
the corresponding stochastic policy after state aliasing ($\pi_{p}$ as defined by equations \ref{eqn:extra-sensing_action_prob} and \ref{eqn:stochasticized_policy}). We then determine the value of this stochastic policy by equation \ref{eqn:oneHot_policy_value}. Then (in the hill climbing step) for each possible policy change we compute the new policy value, and select the action to change the policy. This is repeated until no better changes can be made. Each step's computational complexity is $O(|S|^4|A|)$; each step tests a number of changes no more than $|S||A|$, and the value of a fixed policy can be computed in $O(|S|^3)$ by computing the state transition likelihoods for that policy and using the following closed form computation (this is the standard equation for value computation in a Markov Reward Process (MRP) (See \cite{MRP} for more details on Markov processes).

\begin{equation}
    \Vec{v_s} = (I - \gamma*P_{ss'})^{-1}*\Vec{r_s}
    \label{eqn:MRP_value}
\end{equation}
Where $\Vec{v_s}$ is the vector of state values, $P_{ss'}$ is the transition probability matrix for a given policy, and $\Vec{r_s}$ is the vector of expected rewards at each state (which can be computed for a fixed policy). 

The total time taken for HAPI will naturally be problem specific; the number of improvement steps will depend on the initial point and the possible improvements in the domain. Additional random restarts can improve the outcome, as is common in hill-climbing approaches. 


\subsection{HEU Branch-And-Bound Policy Search (H-B\&B)}

HAPI is effective in quickly finding a good policy. However, if one wanted the optimal policy, then the following branch and bound approach --which we will refer to as H-B\&B can be used for smaller state spaces. It can also be used to bound the suboptimality of the policy found by HAPI, which can be used to decide if further iterations of HAPI could be worthwhile or not.

This branch-and-bound searches in policy space by choosing an action for a state at each level in the search tree. We assume the reader is familiar with the basics of branch and bound \cite{brusco2005branch}. At any given point in the policy search, only a partial policy is defined. We need a lowerbound, and an upperbound to determine if the node in the search tree should be expanded. We set the initial lowerbound as the value of the policy output by HAPI. 


We still need a helpful upperbound that accounts for the extra-sensing action. To compute this, we use an MDP relaxation of the POMDP-HUE for a given partial policy. This is done by assuming perfect state observability \emph{only} for the remainder of the undefined state space in $\pi_d$, and bounding the likelihood of errors and extra-sensing actions for the other states. We call this a ``Partially-Controlled MDP" (PC-MDP). We compute the optimal policy (including extra-sensing actions) for this PC-MDP using value iteration and that is the upperbound. This idea of using an easier MDP to bound the state-value in branch-and-bound is similar to the bound employed in \cite{meuleau2013solving} except theirs does not consider or allow any notion of extra-actions. The proof for the upperbound is simple and is given in the supplemental material under the ``Branch and Bound" section. The gist of it is as follows: If one can set a lower-bound for the probability of state-misidentification and extra-sensing actions in all states, then by optimizing for the remainder of the action probability in each state, the policy value obtained will be equal to or greater than any other possible policy completion. A trivial lower-bound would be to assign zero probability to errors, i.e. $s_1 \neq s_2 \rightarrow p_c(s_1|s_2) = 0$, and extra-sensing actions ($\pi_{p}(s,a^{+}|\pi_d) = 0$) for the states whose policy is not yet defined. Then optimizing the PC-MDP policy would give the upperbound for state-value. Our bound considers the effect of prior decisions in H-B\&B to give a tighter, more helpful upperbound for the search process by using the human model $H$ parameters and assuming undefined states will have the same action to minimize likelihood of extra-sensing action. The pruning effects of our upperbound will be shown in the results, and more details on the bound is in the supplemental section. 

In each step in the branch and bound, we need to run value iteration on the PC-MDP. In our algorithm, we stop value iteration after a certain number of iterations; we set this to 1000 in our experiments. We then take an upperbound for each state's value computed as $v_k(s) + \frac{\epsilon*\gamma}{1-\gamma} $(Chapter 17 \cite{russell2021artificial}) where $\epsilon$ here is $||v_k-v_{k-1}||$. Alternatively, one could set a target error $\epsilon$ and determine the number of iterations needed as $\lceil \frac{log(2R_{max}/(\epsilon(1-\gamma))}{log(1/\gamma}\rceil$ \cite{russell2021artificial}. This error is added to the policy value to set the upperbound. 

The size of the policy search tree is unfortunately large --$|A|^{|S|}$ if we assume the same number of actions (|A|) in each state-- but a good upperbound and ordering the states intelligently can greatly prune the tree. The score, or heuristic we use to order the states is:
\begin{equation}
    \begin{split}
    score(s) = (\frac{1}{|S|}+p_i(s))\times  \sum_{s' in S}p_c(s|s') \times \max_{a \in A(s')}r(s',a)  
    \end{split}
    \label{eqn:bnb_ordering_heuristic}
\end{equation}

This function increases the score of a state based on how likely a state is to be the initial state ($p_i(s)$) since those state values determine the overall policy value (Equation \ref{eqn:oneHot_policy_value}). It also considers the likelihood that other states are confused with it, because then it's policy decision will affect the state value in other states too. This likelihood is scaled by the max reward possible in the other states. The scaling is because we want to prioritize states that can have an influence on the policy in high reward states. This score can help us make pivotal policy decisions sooner in the search process, and work with the upperbound to prune the search tree faster.


\section{Experiments and Results}

We tested our algorithms on two qualitatively different domains; gridworld and warehouse-worker. The warehouse-worker is a domain in which the agent has to make packing decisions for a set of products to be shipped to a customer. The errors come from not knowing which the best box is for a set of products (small, medium or large boxes). We present the gridworld experimental results here, and the warehouse-worker results in the supplemental material along with a more detailed description of the warehouse packer domain.


In our experiments, we repeated HAPI ten times (10 restarts) for each setting and consider the best value as the output from HAPI. H-B\&B ofcourse need only be run once. All results can be consistently reproduced from our codebase all variability in the program is controlled by a random-seed parameter. This is set to 0.

\subsection{ Experimental Setup}


First we present the Gridworld experiments on $5\times5$ grids where H-B\&B was allowed to run to completion. We varied the properties of the MDP to see how well HAPI performs compared to  H-B\&B. The actions for each state are the standard moving up, down, left, and right. The goal is the bottom right square like in Figure \ref{fig:policies_5x5}, which is an absorbing state. The reward is 100 upon transitioning into it. The extra-sensing action ($a^+$) reduces by half both the error likelihood in $p_c$ and probability of incorrect possible-sets $p_u$ ( for possible-sets that have states other than the ground truth). As defined earlier, taking $a^+$ results in a state transition to a parallel state with the same action effects (recall Figure \ref{fig:extra_policy_mental_state}) but with the updated $p_c$ and $p_u$ functions. Subsequent extra-sensing actions from this updated state, return to the same state. This means subsequent repetitions of the extra-sensing action in a state does not affect the inference outcomes ($p_c$ and $p_u$) in our experiments. 

All action transitions are stochastic with a $5\%$ chance of the transition from a random action. All action have a base cost of 0 , but we will add random values to these for each experimental setting. An invalid action (like moving up from the top of the grid) results in the agent staying in the same state and incur the random cost assigned to that action. The extra-sensing action has a cost of 1.

As for the likelihoods of confusing states ($p_c$), we define the likelihood of confusing a grid state (position) with another based on the L1 distance as defined in Equation \ref{eqn:gridworld_classification}. The equation is simply saying that neighboring states are much more likely to be confused with a state than those further away.
\begin{equation}
    p_c(s|s') = \frac{1/(L1(s,s')+\mathbb{1}[s=s'])^m}{\sum_{s'' \in S}1/(L1(s,s'')+\mathbb{1}[s=s''])^m}
    \label{eqn:gridworld_classification}
\end{equation}

where $m$ is a scalar. We set it to 5 for our experiments. This makes the likelihood of confusing with a state more than 1 step away to become very small. Lastly, we add the $+1$ to avoid dividing by zero. As for the possible-sets in $p_u$, we limit ourselves to sets of size 1 and 2. The probability of each set ($p_u(.)$) are computed using equation \ref{eqn:gridworld_uncertainty}. 



\begin{equation}
    \begin{split}
        p_u(\{s_2,s_1\}|s*) =p_c(\hat{s_2}|s_1) *p_c(\hat{s_1}|s*) + \\ p_c(\hat{s_1}|s_2) *p_c(\hat{s_2}|s*) 
   \end{split}
    \label{eqn:gridworld_uncertainty}
\end{equation}

Note $s_1$ can be the same as $s_2$ in Equation \ref{eqn:gridworld_uncertainty}; those cases correspond to the $p_u(.)$ probability for possible-sets of size 1. Lastly, the bias term $\psi_0$ was set to 0.05, which means that even when the agent's possible-set of states has the same policy action, they will take the extra policy action 5\% of the time. We set the extra-sensing likelihood scaling-ratio $\psi_1$ to 0.9.  

We first present results of a 5x5 grid, with one additional mental-state per state. We used a smaller grid first because while HAPI (hill-climbing) can handle larger sizes of grids, branch and bound (H-B\&B) speed drops very quickly as the policy space grows as $|A|^{|S|}$; Even a 5x5 grid has $4^{25} \approx 1.1 \times 10{15}$ policies. However, most actions are quickly eliminated, and the bound helps immensely with pruning the policy space (as results will show). Engineering improvements for parallelization and memory management for H-B\&B is out of the scope of this paper. We posit that for a single task's policy (for a human agent) even state-sizes in the vicinity of 25 can be useful for some problems. For example, a car-maintenance policy for owners would have a few states and associated actions to deal with oil-change, car battery-health and such. In the supplemental section, we discuss another application example in the warehouse-worker domain that can be handled with fewer states. 




To see the performance of the algorithms, we focused on varying three parameters: (1) The discount factor $\gamma$, whose baseline value is 0.7; (2) the likelihood of random action whose default value is $\rho = 0.05$; (3) A ``reward noise range" parameter (RNR) to add random rewards to each of the actions in the grid and whose default value is 2; an $RNR=2$ would result in random additional rewards for each action in the range $[-1,1]$, i.e., uniformly distributed about 0. 

We chose to vary the discount factor since a larger discount factor couples the policy decisions more strongly (since the state value is affected by states further away). We also chose to add random rewards to each of the actions other than the goal state actions to make the search more challenging. Lastly, increased random action likelihood meant that states which had both high reward and high cost actions are less attractive than if there was no random action likelihood. 

Our code was implemented in python using ``pybnb" library for branch and bound, and PyTorch and NumPy for matrix operations. The experiments were run on a PC with Intel® Core™ i7-6700 CPU, running at 3.40GHz on Ubuntu 20.04 with 32 GB of memory.

\subsection{Results}

We first present the values of the policies discovered for the 5x5 grid experiments in Table \ref{tbl: 5x5_policy_values}. The best policy discovered by HAPI approach was either very close or optimal in all cases. The takeaway is that for this problem setting, HAPI with 10 iterations found either a very competitive value or the optimal policy value in all cases. We found this to be the case in the 10x10 grid setting as well in Table \ref{tbl: 10x10_policy_values}. For those experiments, we only used H-B\&B to find an upperbound and define the suboptimality of the policy value found by HAPI. In the 10x10 grid setting as well, HAPI was able to perform well. A corollary is that H-B\&B was able to find a good value-upperbound within 30 minutes, which can be helpful in deciding if further iterations of HAPI might be worthwhile or not. In our supplemental material, we present the timing results for the same experiments, as well as the number of nodes openend by H-B\&B search which shows the efficacy of the upperbound used in the search.

\begin{table}[H]
\scriptsize
\centering
\resizebox{\columnwidth}{!}{%
\begin{tabular}{|l|l|l|l|l|l|}
\hline
\multicolumn{2}{|c|}{\textbf{\begin{tabular}[c]{@{}c@{}}Discount Factor\\ (RNR=2, $\boldsymbol\rho$=0.05)\end{tabular}}} 
& \multicolumn{2}{c|}{\textbf{\begin{tabular}[c]{@{}c@{}}Reward Noise Range\\ ($\boldsymbol\rho$=0.05, $\boldsymbol\gamma$=0.7)\end{tabular}}} 
& \multicolumn{2}{c|}{\textbf{\begin{tabular}[c]{@{}c@{}}Random Action Probability\\ (RNR=2, $\boldsymbol\gamma$=0.7)\end{tabular}}}\\ \hline

$\boldsymbol\gamma$ & Values & RNR & Values & $\boldsymbol\rho$ & Values \\ \hline

0.3 & 11.87,11.87,1.0 & 0 & 33.67,33.67,1.0 & 0.05 & 34.16,34.17,1.0 \\ \hline 
0.5 & 19.05,19.05,1.0 & 1 & 33.89,33.89,1.0 & 0.1 & 33.65,33.65,1.0 \\ \hline 
0.7 & 34.16,34.17,1.0 & 2 & 34.16,34.17,1.0 & 0.15 & 33.08,33.08,1.0 \\ \hline 
0.9 & 69.45,69.45,1.0 & 4 & 34.75,34.75,1.0 & 0.2 & 32.46,32.46,1.0 \\ \hline 

\end{tabular}}
\caption{Policy value results for a 5x5 grid; each entry has the best policy value from HAPI, the optimal value found by H-B\&B, and the ratio of the two. All values rounded down to two decimal places
}
\label{tbl: 5x5_policy_values}
\end{table}

\begin{table}[H]
\scriptsize
\centering
\resizebox{\columnwidth}{!}{%
\begin{tabular}{|l|l|l|l|l|l|}
\hline
\multicolumn{2}{|c|}{\textbf{\begin{tabular}[c]{@{}c@{}}Discount Factor\\ (RNR=2, $\boldsymbol\rho$=0.05)\end{tabular}}} 
& \multicolumn{2}{c|}{\textbf{\begin{tabular}[c]{@{}c@{}}Reward Noise Range\\ ($\boldsymbol\rho$=0.05, $\boldsymbol\gamma$=0.7)\end{tabular}}} 
& \multicolumn{2}{c|}{\textbf{\begin{tabular}[c]{@{}c@{}}Random Action Probability\\ (RNR=2, $\boldsymbol\gamma$=0.7)\end{tabular}}}\\ \hline

$\boldsymbol\gamma$ & Values & RNR & Values & $\boldsymbol\rho$ & Values \\ \hline

0.3 & 3.17,3.48,0.91 & 0 & 11.14,11.47,0.97 & 0.05 & 11.56,12.29,0.94 \\ \hline 
0.5 & 5.26,5.65,0.93 & 1 & 11.27,11.83,0.95 & 0.1 & 11.17,11.86,0.94 \\ \hline 
0.7 & 11.56,12.29,0.94 & 2 & 11.56,12.29,0.94 & 0.15 & 10.77,11.4,0.94 \\ \hline 
0.9 & 43.9,45.91,0.96 & 4 & 12.28,13.48,0.91 & 0.2 & 10.36,10.97,0.94 \\ \hline 

\end{tabular}}
\caption{Policy value results for a 10x10 grid; each entry has the best policy value from HAPI, the best upperbound found by H-B\&B after 30 minutes, and the ratio of the two. All values rounded down to two decimal places}
\label{tbl: 10x10_policy_values}
\end{table}

\subsection{Human Subject Experiments}



\begin{figure}[H]
\centering
     \includegraphics[width=\columnwidth,scale=0.5]{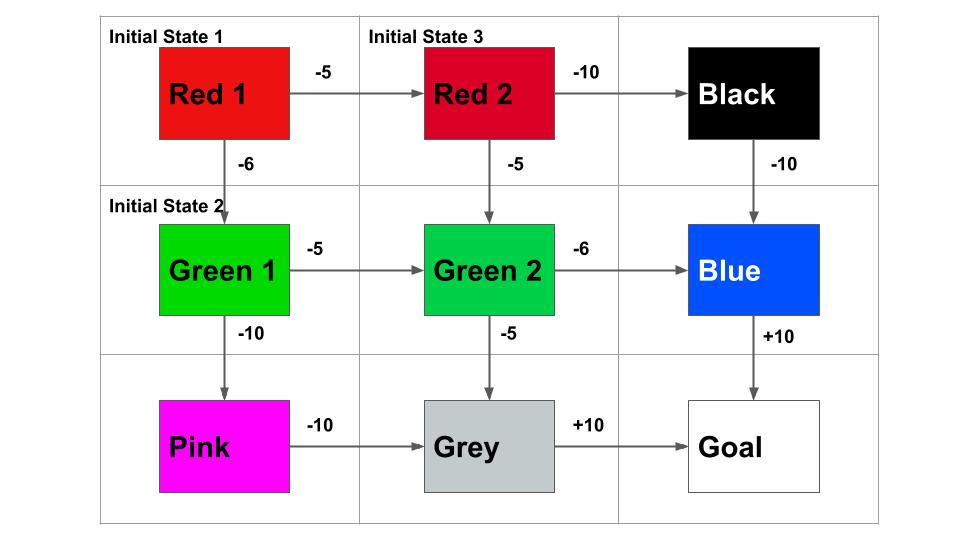}
      \caption{ Policies with identical value in an MDP, but can have different values after accounting for errors and uncertainty effects}
      \label{fig:human_studies_color_grid}
\end{figure}

For our human subject experiments we wanted to see if our assumptions hold with respect to human policy execution under state uncertainty. We use a small grid world setting, and use colors to define each state in order to introduce state uncertainty from perception. The full grid is as illustrated in Figure \ref{fig:human_studies_color_grid}. In the underlying MDP, each action has costs as illustrated in Figure \ref{fig:human_studies_color_grid}, and all undisplayed actions have a cost of -10. There is a reward for reaching the bottom right position (+10). After reaching the goal position, the state then changes to a random new position from a set of initial states.

Our objective in this study was to first build an averaged human model using the procedure described in the paper, and then use that model to compute a policy that accounts for the uncertainty effects. The performance using this policy is then compared to the human agents' performance using the optimal policy. The state space was designed so some sets of states (the color associated to them) are visually similar, like ``Green 1" and ``Green 2" in Figure \ref{fig:human_studies_color_grid}, and cause uncertainty.

All participants were recruited using the ``Prolific" (www.prolific.co) service for online studies, and prescreened using using their service for vision (can see colors clearly). We also asked 3 questions at the start of our study to test if participants could distinguish between lighter and darker shades that look similar. All prolific participants are above 18 years of age, and equal division of male and female participants (as they identify) was requested for the study (Prolific handles this part). No other demographic information was collected. Gender or age based comparisons are out of the scope of this study. Our human-subject studies had IRB approval, and we gave clear information about the purpose of the experiments to our participants in the consent form before the experiments, as well as a debrief message, and the option to contact us for more information.

In the first phase, we collect data to build an (averaged) model of a human agent for the task. This means computing $<p_c, p_u,\psi_0,\psi_1>$ for each state (including the additional mental states). We do this using a preliminary study that displays the same colors as the main study, and asks the human to match a color displayed (colored square) on the left of the screen to a list of numbered colors on the right of the screen. 

The color was also only shown for 0.5 seconds, but the table on the right was permanently displayed. They were given the option to see-again if they were uncertain by pressing the back arrow key; this gives us a clear indication of the extra-sensing action. In this study, before the human can submit their answer or ask to see the color again, we ask them to enter their guesses as to which colors they think it could have been. Example, if a color $Green1$ was displayed the user might enter $(1,2)$ which are the indices for the two green shades, or just enter one color if they were confident in their decision. We give an addition $0.1$ dollars for every correct answer as an incentive to get it correct. We used data from 16 participants for this phase, each participants was asked 15 questions; colors were randomly sampled during testing. We use the data collected to compute the human model as per the equations in the Human Agent Model section, including for the additional mental state after the extra-sensing action. 

As one might expect, the two shades of green, and two shades of red causes participants to request to see the color again, as well as make the most mistakes. All the unique (non-similar) colors that we tested the participants with were easily identified with almost no errors and no see-again actions. We saw this for many unique colors during our initial testing, and so felt confident that we were not showing the colors for too short a duration. One interesting note was that people seemed more likely to confuse the darker shade of red with the lighter shade than vice versa, and this was not the case for the shades of green which was much less confusing. Another interesting point is that $\psi_1$ was higher for the 3 of the confusing colors (Green1,Green2,Red1) after the extra-sensing action was taken. As in they are more likely to take additional extra-sensing actions for those states after the first one. This might be because once a person is uncertain, they tend to doubt their inference more; we do not speculate more on this.

\begin{figure}[H]
\centering
     \includegraphics[width=\columnwidth,scale=0.5]{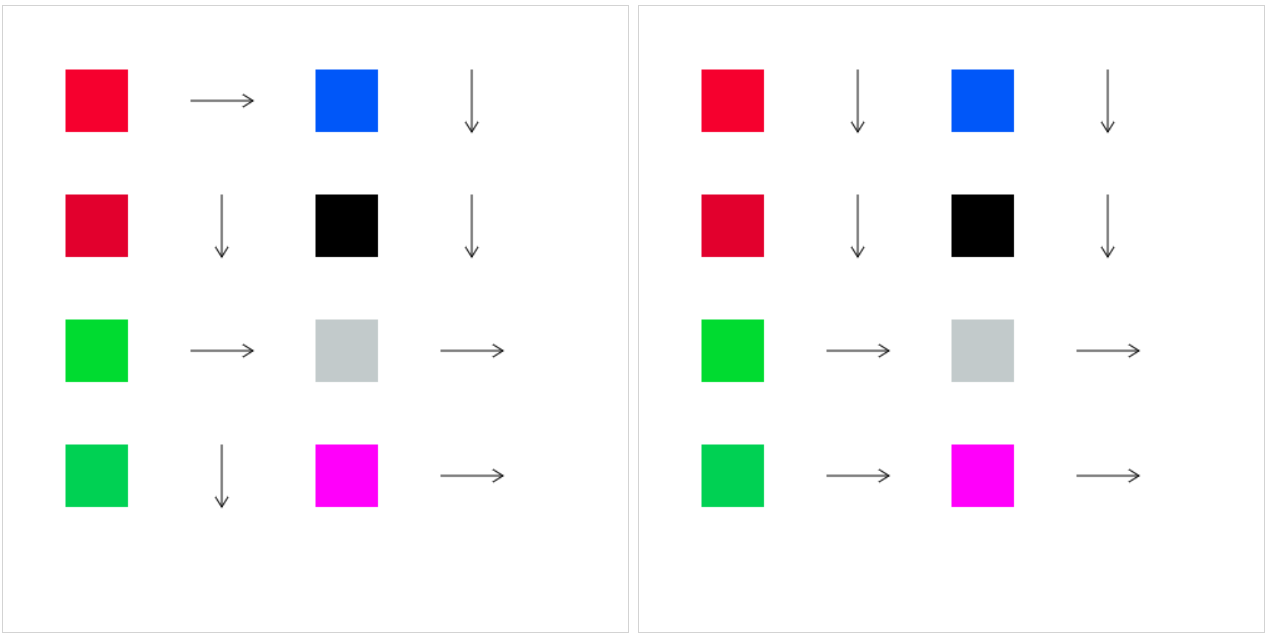}
      \caption{ The two policies for the second phase of human subject experiments. The left policy is the optimal policy after accounting for uncertainty, and the right is the optimal policy without considering the effects of uncertainty. }
      \label{fig:human_studies_phase_2}
\end{figure}

With the data collected, we computed the human-model parameters using the equations defined previously. We then used it to compute the optimal policy using H-B\&B for the grid MDP in the study. We also computed the optimal policy by value iteration which ignores state-uncertainty. These were the two policies given to the humans in phase 2 of the study, and are displayed in Figure \ref{fig:human_studies_phase_2}; the policy on the left is the one that accounts for uncertainty, and the policy on the right is the optimal policy of the MDP, ignoring uncertainty.




Just as in phase 1, we display a color on the left of the screen for 0.5 seconds. This color corresponds to their current position in the grid as in Figure \ref{fig:human_studies_color_grid}. We ask the participant to press the arrow key corresponding to the color seen and so navigate through the underlying grid. After each step the color of the new position is displayed. The participant can see a color again by pressing the "Control" button. We track every button press and their progress through the grid. When they reach the goal state, the state is reset to another of the initial states, for a total of 3 times. The initial states are shown in Figure \ref{fig:human_studies_color_grid}. We used 20 participants for this phase, and had to drop one data point as the data suggested they were randomly guessing for both policies.

Using the data collected, we wanted to see if the reward accrued using the policy that accounted for uncertainty translated to more reward accrued. We took the difference from between the reward accrued in each run (initial to goal state) with it's corresponding run in the other policy. On average the cost incurred by the policy that accounts for uncertainty is less by $1.45$. We ran a dependent t-test for paired samples, as the same participant did both corresponding runs, we treat them as paired samples. We set the significance level at $\alpha = 0.05$, and ran the paired t-test using the scipy-stats library in python \cite{scipy} to see if the policy with uncertainty was better (one-sided). We got a t-test statistic value of $2.240$ which corresponds to a p-value of $0.0147$. Thus we can reject the null hypothesis that the policy with uncertainty is the same or worse than the optimal policy without uncertainty. 

For our experiments, given the simplicity of the problem we found it sufficient and easier to build an averaged model. Ideally one would build a human model unique to a person. In an application, such an averaged model could be tuned with fewer additional data points per person, using a Bayesian approach to the parameters. Overall, our human-subject studies give support to the idea that policies which consider state uncertainty are executed faster and more reliably.

\section{ Related Work}

If the effect of uncertainty was limited to only erroneous state detection (as captured by $p_c(.)$ in the human model), one can frame the problem in this paper as computing the reactive controller for a POMDP and use prior methods in \cite{littman1994memoryless} and \cite{meuleau2013solving}. However, non of the prior methods handle the case where additional (extra-sensing) actions are taken by the agent due to state uncertainty or any function of policy decisions. This can result in different policies between our approach and prior reactive controller approaches if such actions are indeed taken; We verified this by setting $\psi_1,\psi_0$ to be zero (so no extra-policy actions are taken). This reduces our H-B\&B to computing the standard POMDP reactive controller based on $p_c(.)$ as the observation likelihood. We found that the policy and values were suboptimal when there were higher costs of extra-sensing actions; this fits our intuition as the reactive controller is explicitly ignoring those. 

An extension of reactive memoryless controllers is history-based controllers \cite{AkshatKumar2015history} which considers mapping history of observations to policies as opposed to just the current observation (state history of 1). These policies might work well for human agent execution as one need only consider the history of states and not ask the human to explicitly track prior probability distributions, which is a much higher cognitive load. However, such policies are still susceptible to state uncertainty in the human's mind for the same perceptual and cognitive difficulties as in the single observation (single history) case. For this work we limited ourselves to history of 1 state. We will consider extending this to longer histories in future work, and we do not think it a trivial extension as the human model needs to consider sequences of possible-states, and how state-action decisions affect the uncertainty.

On a different but related note, there is work that considers ``blindspots" in an agent's representation \cite{blindspot_safe2sim_ramakrishnan2020blind}. These blindspots can arise due to a mismatch in the state space during training versus execution, or limitations in representational capabilities. There is a follow-up work that focuses specifically on a human agent's blindspots \cite{human_blindspot_ramakrishnan2021bayesian} and reducing errors. Our work attacks the problem from a different angle, and do not try to minimize the errors for a given policy; we instead try to compute a policy that accounts for human errors and behavior in response to uncertainty. 

With respect to errors in state identification, this has also been studied in POMDPs for agents with ``active-perception" capabilities \cite{LionAlg_whitehead1991learning},\cite{CSQL_tan1991cost}, \cite{whitehead1995reinforcement}. In \cite{whitehead1995reinforcement} the authors call the problem as perceptual aliasing when an internal state maps to more than one external state due to limitations of the sensing process. Their approach of "Consistent Representation"(CR) pulls together the prior work on the perceptual aliasing problem. The common assumption across those works is that a consistent representation of the state that is Markovian can be built from the immediate environment with sufficient sensing actions. \cite{whitehead1995reinforcement} also considered "stored-state" architectures where history was incorporated for state inference. These works assume a deterministic agent which is not the case with human agents. We consider how inference can vary after sensing actions, and how that affects policy execution, which is not the case in these works; they assume agent inference given the same information is consistent (mechanistic). 





\section{Conclusion and Future Work}


In this paper, we define the problem of computing reactive policies that account for human execution behavior under state uncertainty. We formalized a probabilistic model of the human agent's inference, and define how to compute the parameters in it. We then present two algorithms (HAPI and H-B\&B) to compute policies that account for behavior under uncertainty, and show experimental results in a gridworld setting. We also have results for a warehouse-worker domain in the supplemental material. Lastly, we conducted human subject studies to show an example of how the human model can be empirically derived, and use it with our H-B\&B algorithm to compute the optimal policy that accounts for uncertainty. We show that this policy resulted in statistically more reward accrued. Our human-subject studies supports the phenomenon we describe in our human model such as errors between similar states due to perception limitaitons, and extra-policy actions when uncertain. 

\bibliography{references}

\include{supplemental}

\end{document}

%% file: supplemental.tex
\appendix

\section{Supplemental Material Contents}

The code for our algorithms and experiments is provided along with the supplemental material, and name references have been removed.

The following  are the additional contents in this supplemental material.

\begin{itemize}


 \item Additional Gridworld Results
 \item H-B\&B Branch And Bound Details
 \item Warehouse Worker Domain and Experiments
 

\end{itemize}





\section{Additional Gridworld Results}

We present in Table \ref{tbl: nodes_opened_bnb} the number of nodes opened by our branch and bound search before finding the optimal solution for 5x5 grid. The number of nodes is significantly less than the policy space of $5^{16}$, which is helpful because the cost of each node is high since we run value iteration on each node for the associated PC-MDP. This also shows that our upper bound was helpful in pruning the policy search tree. 

\begin{table}[H]
\scriptsize
\centering
\resizebox{\columnwidth}{!}{%
\begin{tabular}{|l|l|l|l|l|l|}
\hline
\multicolumn{2}{|c|}{\textbf{\begin{tabular}[c]{@{}c@{}}Discount Factor\\ (RNR=2, $\boldsymbol\rho$=0.05)\end{tabular}}} 
& \multicolumn{2}{c|}{\textbf{\begin{tabular}[c]{@{}c@{}}Reward Noise Range\\ ($\boldsymbol\rho$=0.05, $\boldsymbol\gamma$=0.7)\end{tabular}}} 
& \multicolumn{2}{c|}{\textbf{\begin{tabular}[c]{@{}c@{}}Random Action Probability\\ (RNR=2, $\boldsymbol\gamma$=0.7)\end{tabular}}}\\ \hline

$\boldsymbol\gamma$ & Values & RNR & Values & $\boldsymbol\rho$ & Values \\ \hline

0.3 & 41713 & 0 & 20649 & 0.05 & 7981 \\ \hline
0.5 & 5525 &  1 & 12573 & 0.1 & 7297\\ \hline
0.7 & 7981 &  2 & 7981 & 0.15 & 6785\\ \hline
0.9 & 11933 &  4 & 27589 & 0.2  & 6401\\ \hline
\end{tabular}}
\caption{Number of nodes opened by H-B\&B before finding optimal solution, for a policy search tree of size $4^{25}$ in 5x5 grid experiments}
\label{tbl: nodes_opened_bnb}
\end{table}

The time taken by both algorithms are presented in Tables \ref{tbl: time_taken_5x5} and \ref{tbl: time_taken_10x10}

\begin{table}[H]
\scriptsize
\centering
\resizebox{\columnwidth}{!}{%
\begin{tabular}{|l|l|l|l|l|l|}
\hline
\multicolumn{2}{|c|}{\textbf{\begin{tabular}[c]{@{}c@{}}Discount Factor\\ (RNR=2, $\boldsymbol\rho$=0.05)\end{tabular}}} 
& \multicolumn{2}{c|}{\textbf{\begin{tabular}[c]{@{}c@{}}Reward Noise Range\\ ($\boldsymbol\rho$=0.05, $\boldsymbol\gamma$=0.7)\end{tabular}}} 
& \multicolumn{2}{c|}{\textbf{\begin{tabular}[c]{@{}c@{}}Random Action Probability\\ (RNR=2, $\boldsymbol\gamma$=0.7)\end{tabular}}}\\ \hline

$\boldsymbol\gamma$ & Time(sec) & RNR & Time(sec) & $\boldsymbol\rho$ & Time(sec) \\ \hline

0.3 & 30.09,1788.25, & 0 & 31.51,1175.24, & 0.05 & 31.3,521.97, \\ \hline 
0.5 & 38.73,293.87, & 1 & 33.53,769.04, & 0.1 & 30.08,497.26, \\ \hline 
0.7 & 31.3,521.97, & 2 & 31.3,521.97, & 0.15 & 31.7,498.22, \\ \hline 
0.9 & 34.37,2148.27, & 4 & 29.87,2285.57, & 0.2 & 33.98,482.82, \\ \hline 

\end{tabular}}
\caption{The total time taken (in seconds) by HAPI, followed by the time taken for H-B\&B in 5x5 grid experiment settings}
\label{tbl: time_taken_5x5}
\end{table}

\begin{table}[H]
\scriptsize
\centering
\resizebox{\columnwidth}{!}{%
\begin{tabular}{|l|l|l|l|l|l|}
\hline
\multicolumn{2}{|c|}{\textbf{\begin{tabular}[c]{@{}c@{}}Discount Factor\\ (RNR=2, $\boldsymbol\rho$=0.05)\end{tabular}}} 
& \multicolumn{2}{c|}{\textbf{\begin{tabular}[c]{@{}c@{}}Reward Noise Range\\ ($\boldsymbol\rho$=0.05, $\boldsymbol\gamma$=0.7)\end{tabular}}} 
& \multicolumn{2}{c|}{\textbf{\begin{tabular}[c]{@{}c@{}}Random Action Probability\\ (RNR=2, $\boldsymbol\gamma$=0.7)\end{tabular}}}\\ \hline

$\boldsymbol\gamma$ & Time(sec) & RNR & Time(sec) & $\boldsymbol\rho$ & Time(sec) \\ \hline

0.3 & 13887.44,1800.19, & 0 & 11439.52,1800.36, & 0.05 & 13978.69,1800.52, \\ \hline 
0.5 & 15079.82,1800.31, & 1 & 10434.4,1800.84, & 0.1 & 12784.0,1800.6, \\ \hline 
0.7 & 13978.69,1800.52, & 2 & 13978.69,1800.52, & 0.15 & 10832.57,1801.0, \\ \hline 
0.9 & 11442.28,1802.08, & 4 & 12596.2,1800.67, & 0.2 & 9614.67,1800.86, \\ \hline

\end{tabular}}
\caption{The total time taken (in seconds) by HAPI, followed by the time taken for H-B\&B in 10x10 grid experiment settings. H-B\&B was terminated in 30 minutes and only the tightest upperbound was taken.}
\label{tbl: time_taken_10x10}
\end{table}

\section {H-B\&B Branch And Bound Details}

\subsection{Upperbound For Partial Policy Completions}
Before we go into the details of the branch and bound proof, let us review the equations that define how the stochastic policy --which is actually executed-- is a function of the deterministic policy, the classification probabilities, and uncertainty probabilities.

The first part is the likelihood of extra-sensing action is a function of the state as well as the policy defined over other states, as follows:
\begin{equation}
    \begin{split}
     \pi_{p}(s*,a^{+}|\pi_d) =\psi_0(s*) +  \psi_1(s*) \times \\
     \sum_{S_i \in 2^S} p_u(S_i|s*) \times 1[ 0 < \sum_{s_1,s_2 \in S_i}1[\pi_d(s_1) \neq \pi_d(s_2)] ]
   \end{split}
    \label{eqn:extra-sensing_action_prob}
\end{equation}

The likelihood of normal policy actions (not extra-sensing) are:
\begin{equation}
    \begin{split}
    \pi_{p}(s*,a| \pi_d) = (1-\pi_{p}(s*,a^{+},\pi_d)) \times \\ 
    \sum_{s_i \in S}p_c(\hat{s_i}|s)*1[\pi_d(s_i) = a]
    \end{split}
    \label{eqn:stochasticized_policy}
\end{equation}

The value of a policy is defined as :
\begin{equation}
    V(\pi_d) = \sum_{s \in S}p_i(s)*V_{\pi_{p}}(s)
    \label{eqn:oneHot_policy_value}
\end{equation}

The key idea in the proof is that we can get an upperbound by relaxing the POMDP to have perfect observability, the same assumption made in \cite{meuleau2013solving}. We call this relaxed MDP, a partiall-controlled MDP (PC-MDP) for reasons that will become clear. During policy search in H-B\&B any node is a partial-policy. By partial- policy we mean that only some of the states have an assigned action in $\pi_d$ (the deterministic policy we give to the human). This partial policy needs to be kept in the PC-MDP for it to be a useful bound. The key difference with the prior work \cite{meuleau2013solving} is that we do not use a cross product between the policy graph and MDP (which doubles the state space), \emph{and} we need to  account for the extra-sensing action. In order to do so we leverage the following lemma

{\bf{Lemma: A partial stochastic policy whose action probabilities are the same or lower than those of another partial policy, \emph{for every action} can be completed and optimized to the same or better value as any completion of the other partial policy}}

\begin{figure}[!ht]
\centering
     \includegraphics[width=\columnwidth,scale=0.5]{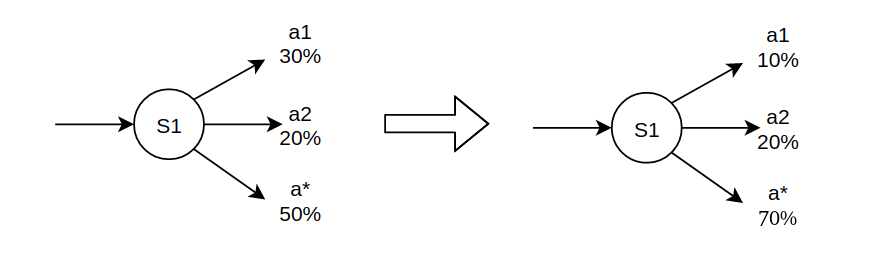}
      \caption{ Example of a partial stochastic policy (left) whose probabilities for every action (a1,a2) are the same or lower than the right policy. ``$a^*$'' is the remaining likelihood that an action can be assigned to (optimizable). }
      \label{fig:lemma_partial_stoch_policy}
\end{figure}

This is best understood starting with an illustration in Figure \ref{fig:lemma_partial_stoch_policy}

\begin{figure}[!ht]
\centering
     \includegraphics[width=\columnwidth,scale=0.5]{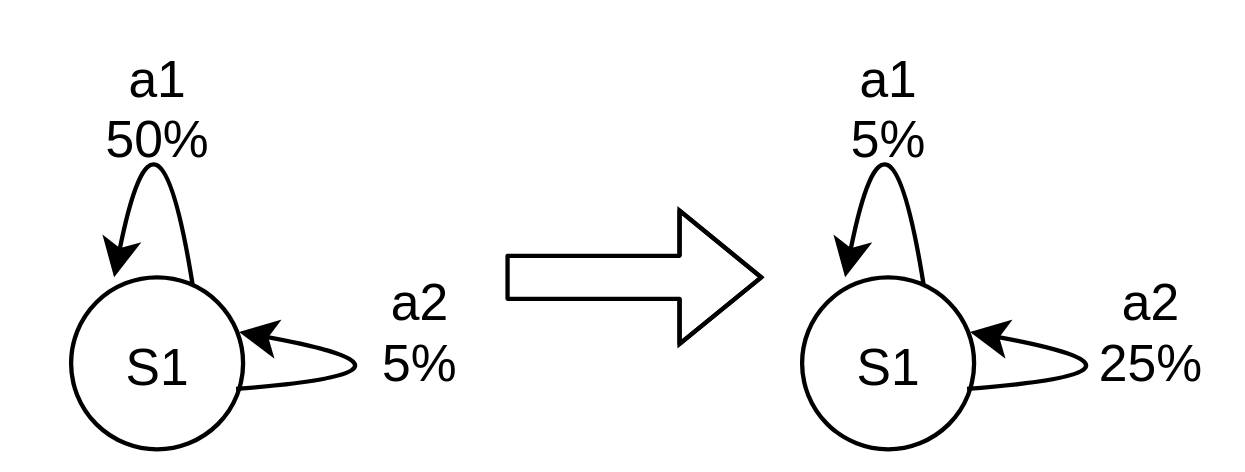}
      \caption{ Example of reduced total probability that will NOT necessarily result in a higher state value after optimization; the lower reward action's probability increases from 5 to 25\%}
      \label{fig:counter_example_reduced_prob}
\end{figure}

Note that it is not enough that the total probability that is optimizable is higher. Even if the overall sum of probabilities for the fixed partial policy goes down, if any one action has increased probability, then this lemma is not guaranteed to hold. A very simple example of this is a single state with two actions that loop back (this is like a multi-arm bandit with discounted future rewards). The first action has a reward of 10, and the second has a reward of 1. If the total percentage of the partially defined policy decreased, but the probability of the second (lower) reward action increases as in Figure \ref{fig:counter_example_reduced_prob}, then we cannot assign more likelihood of the optimal action than in the original partial policy. So the state value would be lower. 

The sketch of the proof for the upperbound value in H-B\&B is rather simple and is as follows. In the PC-MDP, the states can have some of their actions fixed with a certain probability (hence partially controllable). This partial-policy comes from the prior decisions made during policy search in H-B\&B. This PC-MDP can be converted into an equivalent MDP by updating the action transitions and rewards to be the expected sum of each action possible in the state and the actions already defined in the partial policy. The probabilities for this expected sum are determined by the partially defined policy. For this derived MDP, any deterministic policy  will have a value greater than or equal to any stochastic policy in the derived action space (this is a general statement for MDPs)

\begin{figure}[!ht]
\centering
     \includegraphics[width=\columnwidth,scale=0.5]{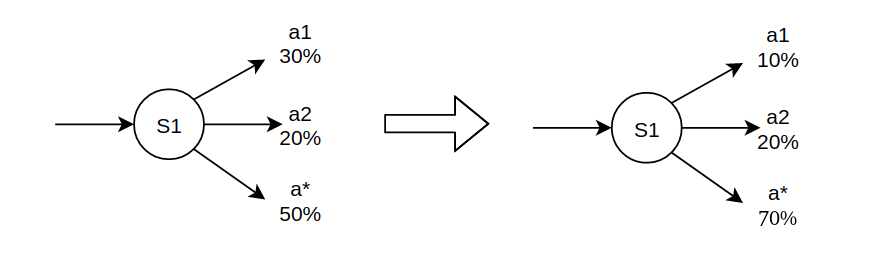}
      \caption{ Example of reduced total probability in a PC-MDP. Left is policy A with 50\% of the action probability accounted for. Right is policy B with 30\% of the probability accounted, and each action's likelihood is lower or equal to the corresponding action in policy A. Note that a* is the placeholder for the optimizable probability amount}
      \label{fig:pcmdp_decreasing_probabilities}
\end{figure}

Now let's connect this to the case that the probability of actions decrease in the partially defined policy. An example of this is illustrated in Figure \ref{fig:pcmdp_decreasing_probabilities}.

If we optimize the MDP associated to the original set of probabilities (before reducing), that results in a stochastic policy in the MDP, lets call this policy A. This same policy distribution as policy A can be chosen for the case when all the action probabilities of the partial-policy are reduced. Let policy B  So the optimal policy value in the reduced probability case will be atleast as good as the value of policy A. So the value of a state  with the optimal policy after the probabilities of the partial policy are reduced will always be greater than or equal to the optimal policy value for the same state with the original probabilities of the partial policy. 
Having made the generic statement of the upperbound, let's now connect it to the search process of H-B\&B.

\subsection{Proof For Upperbound As Applied In H-B\&B}

As we have seen in equation \ref{eqn:stochasticized_policy}, a determinisitic policy translates to a stochastic policy after accounting for uncertainty effects. Using the same functions, a partial deterministic policy at each node of the H-B\&B search space can be mapped to a partial stochastic policy. If we could do so, then we could relax the setting to a fully observable MDP and compute the optimal value of the PC-MDP as an upperbound using the logic stated previously. 

The problem with what we have said so far is that we \emph{cannot} know the partial probabilities of the actions (including $a_{\O}$) until the entire policy is complete. So instead, we compute a lower bound on final action probabilities in a state. The lowerbound for delay probability ($\pi_{sa}(s,a_{\O},\pi_d) $) is computed by considering that all undefined state policies will \emph{not} conflict with the state-policies defined thus far, i.e. $\pi_d(s_i) == \pi_d(s_j)$ where $s_i$ represents the state's whose policy is defined, and $s_j$ is an undefined state. Then the probability is computed as per equation \ref{eqn:extra-sensing_action_prob}. So now we have lower-bounded the probability of $a^+$, this still leaves the normal policy actions.

For the policy action probabilities, the lower-bound likelihood changed from Equation \ref{eqn:stochasticized_policy} to Equation \ref{eqn:pdmax}. In this, we use the \emph{maximum} probability of extra-sensing actions possible from any policy completion. The maximum possible delay is computed by considering the worst case of policy conflicts; every state that can be misclassified and not yet assigned a policy action will have a different action to the current state. This results in the maximum possible likelihood for extra-sensing action. 

\begin{equation}
    \begin{split}
    \pi_{sa}(s,a,\pi_d) = (1-max\_delay(s,\pi_d)) \times \\ 
    \sum_{s_i \in S}p_c(\hat{s_i}|s)*1[\pi_d(s_i) = a]
    \end{split}
    \label{eqn:pdmax}
\end{equation}

Now we have probabilities for actions in the current partial-policy that are less than or equal to the probabilities for the same actions in any possible policy completion. With these probabilities we can get the value of the associated ``PC-MDP" (defined in section 2.1) by value iteration. As we showed in the lemma, the policy value --when all the fixed action probabilities are lower-- will be greater than or equal to any policy completion on the original partial policy. 

We also empirically verified in our experiments that the bound converges from above to the true policy value of any completion of the partial policy. To ensure that it is an upperbound, we add MDP value-iteration error bounds to the policy value computed. The policy value used after k-iterations of value-iteration becomes $v_k(s) + \frac{\epsilon*\gamma}{1-\gamma} $(Chapter 17 \cite{russell2021artificial}) where $\epsilon$ here is $||v_k-v_{k-1}||$. Alternatively, one could set a target error $\epsilon$ and determine the number of iterations needed as $\lceil \frac{log(2R_{max}/(\epsilon(1-\gamma))}{log(1/\gamma}\rceil$ \cite{russell2021artificial}.

With the error bound added, we have a true upperbound for H-B\&B at every node in the search.



\section{Warehouse Worker Domain and Experiments}

\subsection{Description of the warehouse worker domain}

To help understand and emphasize the value of this problem, as well as the ideas in this paper, let us consider the setting of a packer in a warehouse. In the warehouse-worker setting, the human agent's job is to put a set of products into the best fit box for shipping to the customer. The products come in various shapes and sizes. Any given instance is a set of products will belong to one of the following states: small, medium, and large, which also corresponds to the best action (optimal box size to fit it in).
The human is trained in determining the state (think as classifying the state associated to an instance). Let's say they were trained on a set of prototype examples of small, medium and large orders and with this experience, they will make decisions based on similarity (size, shape) to these examples. They have to decide if a set of products can be arranged to fit into a box of a certain size. This decision is related to the bin packing problem, which in itself is a challenging problem. Needless to say, no manageable set of training prototypes or features to describe them, will completely cover the space of all possible instances; different oddly shaped products maybe stacked efficiently into a small box, but it is hard to know when or how. Some instances (set of products) will lie near the decision boundary between states in the human's mind. This can lead to uncertainty in the mind of the human. This uncertainty is from cognitive limitations (thinking of different arrangements to fit into a box) and/or perceptual limitations (determining relative sizes of products accurately from sight). As a result of this uncertainty, they may delay acting because they are uncertain as to what box to put a product into; they may additionally try orienting the products differently (additional perception action) or ask the supervisor for help. This is what we call a extra-sensing action (not controlled in the execution policy). The specific dynamics of extra-sensing action(s) are domain specific. The defining feature of them, is that they cannot be deterministically controlled in the human agent. After these additional extra-sensing actions, the human may still make a mistake in determining the right state, and so might choose the wrong action as well. 
In our warehouse worker setting, the worker might misclassify a set of products that would fit into a medium-sized box, as a "small-box" state instance, and try to fit them into a smaller box. This would result in some left over products that would not fit. Now the problem state has the left-over products, which the human has to use another box to put the remainder into, or they may choose to repack everything into the larger box. Either way, the cost of such additional actions from a human's erroneous execution can make a seemingly optimal policy that is poorly executed to become woefully suboptimal. 
If, however, we could account for such errors, we can compute policies that reduce them. For example if the extra cost of always using the large box is small in comparison to the cost of errors and delays from a different policy, then a good policy-search algorithm should return a policy that says to always use the large box. There would be no delay actions, or additional perception actions due to uncertainty, leading to more products shipped over time. 

\subsection{Warehouse Worker Domain Setting }

In the warehouse-worker setting, the human agent's job is to put a set of products associated to a customer's order into the best fit box for shipping to the customer. The orders come down a conveyor belt, and the products in the orders come in various shapes and sizes. The states in this domain are comprised of a set of products. What state a set of products belongs to, is determined by the optimal box that the products will fit into, and if bubble wrap is needed. The states and the packing actions map 1-to-1. In a simple instance of the domain, a given set of products will belong to one of the following states: small, medium, and large, which correspond to the optimal box size to fit it in.

The human worker is trained in determining the state (think as classifying the state associated to a set of products). We assume the worker was trained on a set of prototype examples of small, medium and large orders and will make decisions based on similarity (size and shape) to these examples. They have to decide if a set of products can be arranged to fit into a box size. Such a decision rivals in complexity to the bin packing problem, which can be very hard. Needless to say, no manageable set of training prototypes or features to describe them, will completely cover the space of all possible state instances (set of products). Some  state instances will lie near the decision boundary between states in the human's mind, and this can lead to uncertainty. This uncertainty is from cognitive limitations (thinking of different arrangments to fit into a box) and/or perceptual limitations (comparing lengths or sizes of products accurately from sight). Because of this uncertainty, they may try orienting the products differently or ask the supervisor for help. This is what we capture as an extra-sensing action action.

In our warehouse worker setting, the worker might misclassify a set of products as a small-box state, and try to fit a ``medium-box" set of items  into a smaller box. This would result in some left over products that would not fit. Now the state has the left-over products, which the human has to use another box to pack the remaining items. A medium-sized order may still remain a medium sized order after the human failed to fit it into the small-box; this can happen when some of the leftover products are just too wide for the small box. The likelihood of transitions between the different states (sizes of orders) with each packing action is given by the domain designer. 
 
In our conceptualization of this domain, after a worker goes through basic training in the warehouse, the worker is evaluated by the supervisor to evaluate how well they classify orders, and the kind of confusion/errors they make; put in different words, the worker is evaluated to build the human model $\langle p_c, p_u,\psi_0,\psi_1 \rangle$. Based on this, a policy is developed for the worker by considering the company's average estimates for the money made per order when using different types of packaging (reward specification), and the likelihood of order types. This can be used to compute a better policy for the human workers. 
\emph{If} (for example) in a trivial warehouse-worker setting, the loss in reward is very small for always using the largest box, then a policy that says to always use the larger box might be better that the supposed optimal policy of using the exact right sized box. There would be no extra-sensing actions due to uncertainty, or multiple packing actions because a smaller box was used incorrectly. This would lead to more products completed completed over time. So it would be a higher value policy after accounting for the effects of uncertainty and erroneous execution.

\subsection{Warehouse Worker Experimental Setting }

For our experiments, the state and action sets are defined by the cartesian product of the set of box sizes $\{small,medium,large\}$, and if bubble-wrap (soft padding) is needed $\{wrap,no\_wrap\}$. For example a group of items with glass items could be a small order that requires a small box with bubble-wrap, $small \times wrap$. When a worker sees an order, it is not always apparent what box size is needed. Additionally, due to the diversity of products, the worker has no idea which products actually need bubble wrap or not. For example there maybe tempered (hardened) glass products that do not need bubble-wrap, but the worker might not know this and use bubble-wrap anyway. 

The transitions are straightforward, and easily defined in text. If state is a ``large-box with bubble-wrap", and the agent takes the action to package as a medium-order or smaller, then the agent gets a reward for packaging some of the items, and the state transitions 100\% of the time to the same state of ``large-box with bubble-wrap"; we consider there was atleast one large product that didn't fit. The same is done for medium orders when the action taken is to package as a small-order. If the action taken is to use a larger size box for a smaller order, then the order is completed but with a lower reward than if the correct sized package was used. The reward is also lower if bubble wrap was needed but not used. Once an order is complete (the right sized box was used), the state transitions to the next order based on the probability of initial states, which is uniform likelihood for our experiments (16.66\% for each state). This initial state distribution would corresponds to the statistics of the warehouse company on the distribution of order-size.

Thus far we have only talked of package sizes. The other dimension is whether an order needs bubblewrap (soft packaging material) or not. If an order is large and needs bubble-wrap, then packaging it as medium without bubble wrap gives a lower reward and transitions to ``medium-size with-bubble wrap" state 50\% of the time, or medium-size (no bubble wrap) state for the other 50\% of the time. This is to reflect that the user might have packaged the items that required bubble wrap already. The analogous transition happens with medium-sized orders that require bubble wrap.

As for the reward. The right action--right size and addition of bubble wrap if needed-- gives a reward of 1.0. All other suboptimal actions gives a lower reward. How much lower is randomly determined by the Reward Noise Range (RNR) parameter. We vary this parameter between ${0.1,0.2,0.3,0.5}$. Based on this parameter the reward is randomly reduced by upto this amount for each action. 

The other part of the experimental setup is the classification likelihoods which are shown in Table \ref{tbl:ww_classification_matrix}. The likelihood of a worker confusing any small order with any medium sized order or vice versa is about $16\%$, and is the same for misclassifying any medium with any large order. The likelihood of confusing a small order with a large order is less than $1\%$. The likelihood of the worker correctly determining if bubblewrap is needed is $50\%$ across all order sizes; there are so many products that their accuracy for determining if bubble wrap is needed is random. 

\begin{table}[h!]
\scriptsize
\centering
\begin{tabular}{|c|c|c|c|c|c|c|}
\hline
 & l & l $\times$ w & m & m $\times$ w & s & s $\times$ w \\
\hline
l & 32.68\% & 32.68\% & 12.50\% & 12.50\% & 0.98\% & 0.98\% \\
\hline
l $\times$ w & 32.68\% & 32.68\% & 12.50\% & 12.50\% & 0.98\% & 0.98\% \\
\hline
m & 16.34\% & 16.34\% & 25.00\% & 25.00\% & 16.34\% & 16.34\% \\
\hline
m $\times$ w & 16.34\% & 16.34\% & 25.00\% & 25.00\% & 16.34\% & 16.34\% \\
\hline
s & 0.98\% & 0.98\% & 12.50\% & 12.50\% & 32.68\% & 32.68\% \\
\hline
s $\times$ w & 0.98\% & 0.98\% & 12.50\% & 12.50\% & 32.68\% & 32.68\% \\
\hline
\end{tabular}
\caption{Classification Likelihood Matrix ($p_c$) for Warehouse-Worker Domain, where (s,m,l) stands for (small,medium,large) and ``w" means bubblewrap needed.}
\label{tbl:ww_classification_matrix}
\end{table}

The probability of uncertainty $p_u(.)$ is computed using the same method as for the gridworld experiments. The $p_u(.)$ probabilities were set by taking the average of the classification probabilities from each pair of states, computed as :

\begin{equation}
    \begin{split}
        p_u(\{s_2,s_1\}|s*) = (\frac{p_c(\hat{s2}|s*) + p_c(\hat{s_2}|s1)}{2})*p_c({\hat{s1}|s^*})\\
        + (\frac{p_c(\hat{s1}|s*) + p_c(\hat{s_1}|s2)}{2})*p_c({\hat{s2}|s^*})
   \end{split}
    \label{eqn:warehouse_uncertainty}
\end{equation}
We ensured that the probability of the cases for any given state sum to 1. This equation 

This would mean that if two states had a high probability in $p_c$ of being inferred in the state $s*$, then the likelihood that the agent would be uncertain between those two states is higher too. Lastly, we set $\psi_1 = 1.0,\psi_0 = 0$ for our experiments. The extra-sensing action cost is -0.1. The extra-sensing action in our experiments does not improve the inference likelihoods. This represents the case when the human repeats unhelpful sensing actions, or is just delayed due to uncertainty. 

\begin{figure}[H]
\centering
     \includegraphics[width=\columnwidth,scale=0.5]{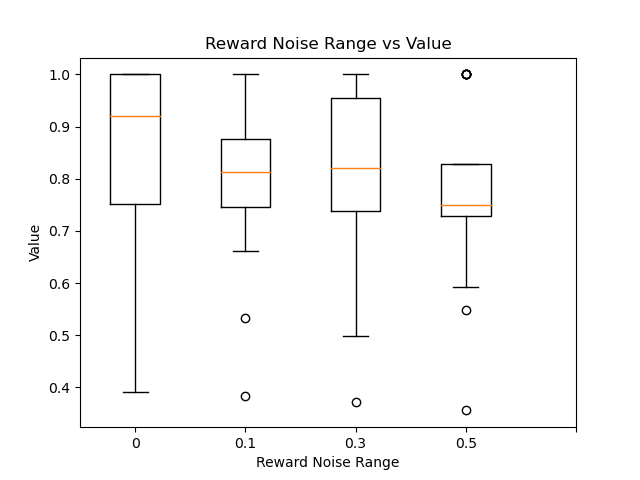}
      \caption{Box plot of HAPI policy values for warehouse-worker domain with varying Reward Noise Range; values normalized by the policy value from Branch and Bound search}
      \label{fig:warehouse_varying_rr_value}
\end{figure}

\begin{figure}[H]
\centering
     \includegraphics[width=\columnwidth,scale=0.5]{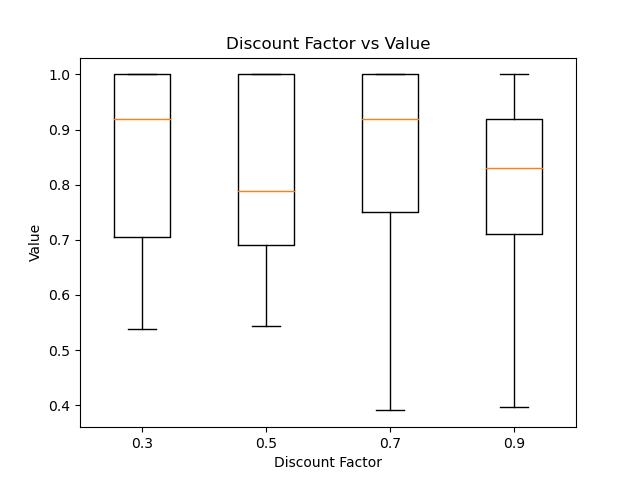}
      \caption{Box plot of HAPI policy values for warehouse-worker domain with varying discount factor ($\gamma$); values normalized by the policy value from Branch and Bound search}
      \label{fig:warehouse_varying_gamma_value}
\end{figure}

In our experiments we vary the reward noise range (RNR) parameter as ${0.1,0.2,0.3,0.5}$ for one set of experiments, these are presented in Figure \ref{fig:warehouse_varying_rr_value}. We also vary the discount factor as ${0.3,0.5,0.7,0.9}$. These results are presented in \ref{fig:warehouse_varying_gamma_value}. The default RNR is 0.0 when it is not being varied, and the default discount factor is 0.7. We also set the probability of taking a random action is 5\% to add more stochasticity into the domain. 
For each setting we run HAPI 30 times and present the distribution of the policy value (Equation \ref{eqn:oneHot_policy_value}) normalized by the value of the policy returned by branch and bound. Since the branch and bound policy value is optimal , we expect the range to be [0,1].

\subsection{Warehouse Worker Experimental Results}

In all the experiments, as we wanted, the branch and bound policy has the highest value. The HAPI policy search found the optimal policy value atleast once out of 30 iterations, which is what we hoped. The variance in policy value found by HAPI is a lot more than in the grid world experiments. The trend of incresing variance in policy value for HAPI as discount factor increases is not as evident in this domain as it was in gridworld. This maybe because of fewer states. 

The surprising trend is that as the RNR parameter was increased (less reward for actions) HAPI seemed to find the optimal policy a lot more consistently. It is easy to see why this holds in the limit when all the rewards are zero and the only cost is from inaction (extra-sensing action). Then the optimal policy is any policy that has the same action across all states, of which there are many and easy to find. We were surprised that the effect of the extra-sensing action was pronounced even with smaller values of RNR. Indeed the optimal policies found were just to put all state orders in the large-box with bubblewrap, which fits out intuition; if the difference in rewards is negligible, then use the action that applies to all states and avoids extra-sensing actions.



